\pdfoutput=1
\documentclass[11pt]{article}

\PassOptionsToPackage{table}{xcolor}

\usepackage[preprint]{acl}   % preprint: non-anonymous, page numbers, no line numbers
\usepackage{times}
\usepackage{latexsym}
\usepackage[T1]{fontenc}
\usepackage[utf8]{inputenc}
\usepackage{microtype}
\usepackage{graphicx}
\usepackage{booktabs}
\usepackage{enumitem}
\usepackage{amsmath,amssymb}
\usepackage{amsthm}
\usepackage{algorithm}
\usepackage{algpseudocode}
\usepackage[dvipsnames]{xcolor}
\usepackage{multirow}
\usepackage{array}
\usepackage{makecell}
\usepackage{tabularx}
\usepackage{pifont}
\usepackage{tikz}
\usetikzlibrary{arrows.meta,positioning,shapes.geometric,calc}
\usepackage{needspace}
\definecolor{gainc}{RGB}{0,140,70}
\definecolor{dropc}{RGB}{200,30,30}
\newcommand{\up}[1]{\raisebox{0.5ex}{\textcolor{gainc}{\scriptsize$\uparrow$#1}}}
\newcommand{\down}[1]{\raisebox{0.5ex}{\textcolor{dropc}{\scriptsize$\downarrow$#1}}}

\definecolor{tableheadcolor}{RGB}{181, 82, 62}   % header tint
\definecolor{oursrowcolor}{RGB}{217, 227, 241}   % our-method rows
\newcommand{\headrow}{\rowcolor{tableheadcolor!20}}

\newcommand{\altrow}{\rowcolor{gray!8}}

\definecolor{caseqfill}{HTML}{F5F4F1}   \definecolor{caseqline}{HTML}{B6B4AE}
\definecolor{stgAfill}{HTML}{FAF0EA}    \definecolor{stgAline}{HTML}{C08A72}
\definecolor{stgBfill}{HTML}{ECF2F8}    \definecolor{stgBline}{HTML}{7FA6C2}
\definecolor{stgCfill}{HTML}{FBEFEF}    \definecolor{stgCline}{HTML}{CE9090}
\definecolor{markgood}{HTML}{2E7D4F}    \definecolor{markbad}{HTML}{B03A2E}

\newcommand{\okmark}{\textcolor{markgood}{\ding{51}}}
\newcommand{\xmark}{\textcolor{markbad}{\ding{55}}}

\newcommand{\casenote}[3]{%
  \par\nointerlineskip\vskip 2.2pt\noindent
  \tikz{\node[inner xsep=4pt,inner ysep=2.4pt,fill=#1,draw=#2,line width=0.4pt,
              rounded corners=2.5pt,align=left,font=\scriptsize,
              text width=\dimexpr\linewidth-10pt\relax] {#3};}%
  \par\nointerlineskip}

\newcommand{\casenoteoff}[3]{%
  \par\nointerlineskip\vskip 2.2pt\noindent
  \tikz{\node[inner xsep=4pt,inner ysep=2.4pt,fill=#1!55!white,draw=#2!75!white,
              dash pattern=on 1.7pt off 1.5pt,line width=0.4pt,
              rounded corners=2.5pt,align=left,font=\scriptsize,
              text width=\dimexpr\linewidth-10pt\relax] {#3};}%
  \par\nointerlineskip}

\newlength{\casenumw}
\newcommand{\casetab}[1]{%
  {\scriptsize\setlength{\tabcolsep}{2.2pt}%
  \setlength{\casenumw}{\dimexpr(\linewidth-1.6cm-12\tabcolsep)/5\relax}%
  \begin{tabular}{p{1.6cm} *{5}{>{\raggedleft\arraybackslash}p{\casenumw}}}
  \Xhline{1.0pt}
  \headrow Stage & A & B & C & D & $m$ \\
  \Xhline{1.0pt}
  #1
  \Xhline{1.0pt}
  \end{tabular}}}

\newsavebox{\caseimgbox}
\newlength{\caseimght}
\newcommand{\casehdr}[4]{%
  \sbox{\caseimgbox}{\includegraphics[width=0.33\linewidth]{#1}}%
  \setlength{\caseimght}{\dimexpr\ht\caseimgbox+\dp\caseimgbox\relax}%
  \noindent
  \begin{minipage}[c]{0.33\linewidth}\centering
    \usebox{\caseimgbox}
  \end{minipage}\hfill
  \begin{minipage}[c]{0.645\linewidth}
    \noindent
    \tikz{\node[inner xsep=4pt,inner ysep=2.4pt,fill=caseqfill,draw=caseqline,
                line width=0.4pt,rounded corners=2.5pt,align=left,
                font=\scriptsize,minimum height=\caseimght,
                text width=\dimexpr\linewidth-10pt\relax]
      {\textbf{Question:}~#2\\[1.5pt]
       \begin{tabular}{@{}l@{\hspace{5pt}}l@{}}#3\end{tabular}\\[1.5pt]
       \textbf{Correct answer:}~#4};}%
  \end{minipage}\par}

\newcommand{\method}{\texttt{MAGIC}}     % rename here if the method name changes
\newcommand{\scores}{S}                  % base score vector
\newcommand{\sbase}{S_0}
\newcommand{\margin}[1]{m(#1)}
\newcommand{\prior}{h}                    % true prior contribution
\newcommand{\est}{\hat{h}}                % prior estimate
\newcommand{\noise}{\varepsilon}
\newcommand{\dpg}{\Delta\prior}            % signed top-two prior gap
\newcommand{\de}{\Delta\noise}            % signed top-two error gap
\newcommand{\Util}{U}

\newcommand{\ton}{t_1}
\newcommand{\ttwo}{t_2}

\newcommand{\tauV}{\tau_{\mathrm{V}}}

\newcommand{\Done}{D_1}
\newcommand{\Dtwo}{D_2}
\newcommand{\Vsrc}{V_{\mathrm{src}}}

\title{\method{}: Margin-gated Contrastive Debiasing with Evidence
Restoration for VLM Test-time Robustness Scaling}

\author{
  Shunshun Gu\textsuperscript{1},
  Enze Zhang\textsuperscript{1},
  Ming Pei\textsuperscript{1},
  Hang Zhou\textsuperscript{2},
  Xiao Luo\textsuperscript{1} \\
  \textsuperscript{1}University of Wisconsin--Madison \quad
  \textsuperscript{2}University of North Carolina at Chapel Hill \\
  \small\texttt{\{sgu82, ezhang62, mpei3, xiao.luo\}@wisc.edu}, \quad
  \small\texttt{hzh@unc.edu}
}

\begin{document}
\maketitle

\begin{abstract}
This paper studies the problem of test-time robustness scaling for large vision-language models (LVLMs), which aims to mitigate language bias during inference. Existing approaches typically adopt test-time scaling techniques such as contrastive decoding to improve inference performance. Despite the progress, the performance of these approaches is far from satisfactory due to intentional evidence removal during perturbations and redundant inference cost. Towards this end, we propose a novel approach named \underline{M}argin-gated Contr\underline{a}stive Debiasin\underline{g} with Ev\underline{i}den\underline{c}e Restoration (\method{}). The core of our \method{} is to evaluate the decision margins of LVLMs to selectively activate textual and visual perturbation branches for efficient debiased inference. In particular, our \method{} first measures the prediction margins of crucial tokens, and applies a gating function to identify low-confidence responses that are susceptible to language biases. For selected samples, we integrate the token logits from multiple textual and image perturbation branches with the original ones for debiased decision. More importantly, we selectively restore the most informative corrupted views with sharp prediction distributions for ambiguous samples after debiasing, enabling evidence recovery for discriminative decision-making. We further utilize a calibration dataset to actively identify and prune ineffective branches during contrastive debiasing for cost reduction. Extensive experiments on benchmark datasets validate the effectiveness of the proposed \method{} in comparison with extensive baselines. 
\end{abstract}

\section{Introduction}
\label{sec:intro}

Large vision--language models (LVLMs) often answer from their language prior
rather than the image \citep{antol2015vqa,goyal2017making,agrawal2018dont}. Asked
about an image whose answer is rare in the world, they tend to return the
common answer regardless of what the image shows \citep{luo2024vilp}.
Correcting this bias at inference time has become an active line of work,
since such correction needs no retraining and applies to any pretrained model.

Most inference-time methods for this problem share a contrastive decoding (CD)
paradigm. They run the model a second time on a corrupted input, such as a
blank or noised image, subtract the corrupted branch's scores from the
original, and take the difference to be the bias. Counterfactual VQA \citep{niu2021counterfactual} subtracts a single image-free
branch, while VCD
\citep{leng2024vcd} and M3ID \citep{favero2024m3id} vary how the input is
corrupted and how the branches are combined. Building on these, SCI
\citep{tang2026sci} averages several perturbation branches and reports that
adding branches yields a stronger correction. Across the family, the
correction is a fixed function of the corrupted branches, applied to every
sample.

However, this family shares two limitations. First, the perturbation
that produces the subtracted branch removes more than the prior. The
correction is built from the corrupted branch alone, so it discards the
image's own contribution along with the language prior. A sample whose answer
turns on that contribution is left without the evidence it needs. Second, the correction is charged to every
input over a branch set fixed in advance. Forward passes are spent on inputs
that a bounded correction cannot change. Each added branch also yields less
than the last, since a perturbed branch estimates the prior rather than being it and
averaging shrinks only part of that estimate's error.

\begin{figure*}[t]\centering
\includegraphics[width=\textwidth]{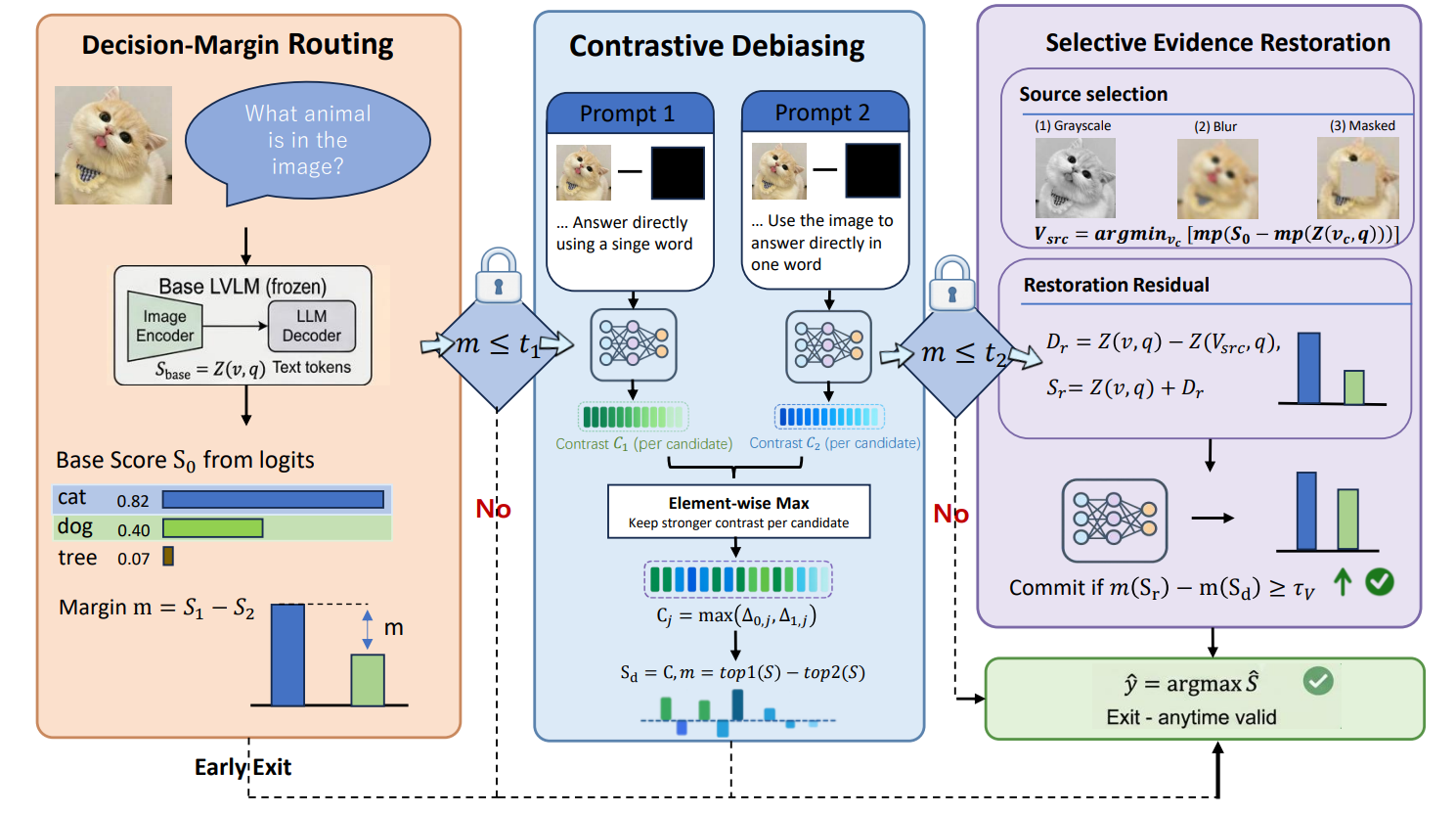}
\caption{Overview of \method. Each step fires only when the margin $m$ still
leaves the decision contested, so a sample that clears a gate exits with the
scores it already has. Thresholds are fitted offline on the validation split.}
\label{fig:teaser}
\vspace{-5mm}
\end{figure*}

To address these limitations, we propose \method, a training-free corrector built
from the three parts shown in Figure~\ref{fig:teaser}. It allocates correction by the
model's decision margin, so confident predictions pass through untouched while
contested ones receive it. It then restores the visual evidence the
perturbation removes, drawn from whichever perturbed view leaves the model
most confident, and keeps it only where the decision becomes more definite.
Finally, it prunes its perturbation branches per model on a calibration set
rather than fixing them in advance. The design carries three thresholds and no trained parameters, and it spends its passes only where the correction is worth its cost. We validate this design with extensive experiments across four LVLM backbones and six datasets, showing that \method\ consistently
outperforms strong contrastive-decoding baselines

Our contributions are summarized as follows. \ding{182} We recast test-time
debiasing as an allocation problem rather than a fixed operation, in which the
extra forward passes a corrector makes are a budget to be spent where it can
change the answer. \ding{183} We propose \method, a corrector that adds no
trained parameters, gates correction on the decision margin, restores the evidence a perturbation removes, and
sizes its branch set per model. \ding{184} Extensive experiments on four LVLMs from
two model generations show that \method\ consistently
outperforms strong CD baselines at a cost that adapts to the input.

\section{Related Work}
\label{sec:related}

\noindent\textbf{Large Vision-Language Models.} LVLMs pair a visual encoder
with a pretrained language model through a lightweight connector, so that
patch-level visual features and text tokens are decoded jointly
\citep{wang2024qwen2vl,liu2024llavanext,bai2025qwen3vl,an2025llavaov15}. Because the language model holds most of
the parameters, decoding is easily
dominated by what the text alone makes likely. This tendency was first studied as
the language prior in visual question answering
\citep{antol2015vqa,goyal2017making,agrawal2018dont}. It persists in current
LVLMs, which return the world-typical answer even when the image shows
otherwise \citep{luo2024vilp}. Encoders, connectors, and training
data differ across releases, so the correction each model needs may
differ as well. Our evaluation therefore spans four backbones from two model
generations.

\noindent\textbf{Test-Time Robustness Scaling.} Rather than retraining, a
frozen LVLM can be made more robust by spending extra forward passes at
inference. Contrastive decoding scores a token by the gap between two
runs of the model, one of them weakened so that the gap suppresses what the
weaker run favors \citep{li2023contrastive,chuang2024dola}. Vision-language
methods weaken the run by corrupting the image and read the gap as the language
prior. VCD \citep{leng2024vcd} noises the image, while M3ID
\citep{favero2024m3id} drops it entirely and strengthens the subtraction as the
response grows.
Counterfactual-inference methods \citep{tang2020unbiased,niu2021counterfactual}
reach the same subtraction through causal effects. SCI
\citep{tang2026sci} averages several such branches. Others make the correction itself adaptive. Some fire
it only where a signal read off the model says it is needed, the predictive
confidence \citep{tang2026ccd,huang2026chasd}, the stability of the token
preference across layers \citep{liu2026htdc}, or the gap between the
multimodal and text-only branches \citep{ren2026nolan}. Others select which
perturbation to contrast against \citep{kim2024vacode} or contrast
iteratively across image scales \citep{park2025second}. In all of them the branch set is
fixed in advance and charged to every sample. Adjacent work acts inside the forward pass
\citep{liu2024pai,huo2025sid,deng2025maskcd,huang2024opera}, at training time
\citep{clark2019dont,cadene2019rubi,utama2020mind}, or on content-free
\citep{zhao2021calibrate,holtzman2021surface} and self-describing
\citep{schick2021selfdiag} prompts. However, subtracting such a branch removes
genuine visual evidence along with the prior. \method\ instead gates the
correction on the decision margin, restores the evidence the perturbation
removes, and prunes its branches per model.

\section{The Proposed \method{}}
\label{sec:method}

\noindent\textbf{Problem Definition.}
\label{sec:estimation}
We consider inference-time correction of a frozen LVLM. Given an image $v$ and
a question $q$, let $Z(v,q)\in\mathbb{R}^{K}$ be the scores the model assigns
to $K$ answer candidates, each scored by the log-probability of its first
answer token. The candidates are the option letters for multiple-choice items
and a pool of the model's own answers for open-ended ones. Writing
$\sbase=Z(v,q)$ for the scores on the real input, the model answers with
$\arg\max\sbase$. On some inputs these scores are carried by the language
prior rather than by the image. Our goal is to correct those inputs without
retraining, while leaving the rest untouched.
Existing correctors subtract from $\sbase$ an estimate $\est$ of the language
prior, which is formulated as follows:
\begin{equation}
\scores = \sbase - \est ,
\label{eq:corr}
\end{equation}
where $\est$ is read from a perturbation branch $Z(\tilde v,\tilde q)$ that
runs the same model on a corrupted input, and existing methods differ only in
how $(\tilde v,\tilde q)$ is chosen. Because this correction is charged to
every input and its branch set is fixed in advance, the budget is spent
uniformly rather than where it can change the answer.

\subsection{Framework Overview}
\label{sec:overview}

In this paper, we propose \method, a training-free corrector that spends the
budget of Eq.~\eqref{eq:corr} only where it can change the answer. The core of
\method\ is to treat correction as a resource allocated by the decision margin
of $\sbase$, so that a confident prediction keeps its single forward pass
while only a contested one is corrected. In particular, \method\ first
replaces $\sbase$ with a state rebuilt from anti-prior contrasts, which
removes the prior. Then, because that removal also discards what the image
contributes, a second gate decides whether to add a visual residual back.
Finally, the branch set both steps draw on is fitted per model rather than
fixed in advance, and no parameter is trained. Zero gates recover the base
model, while an always-open first gate over a single blank branch reduces
\method\ to Eq.~\eqref{eq:corr}. An overview of \method\ can be found in Figure~\ref{fig:teaser}.

\subsection{Debiased Contrastive Decoding with Margin Gating}
\label{sec:gating}

\begin{figure*}[t]\centering
\includegraphics[width=\textwidth]{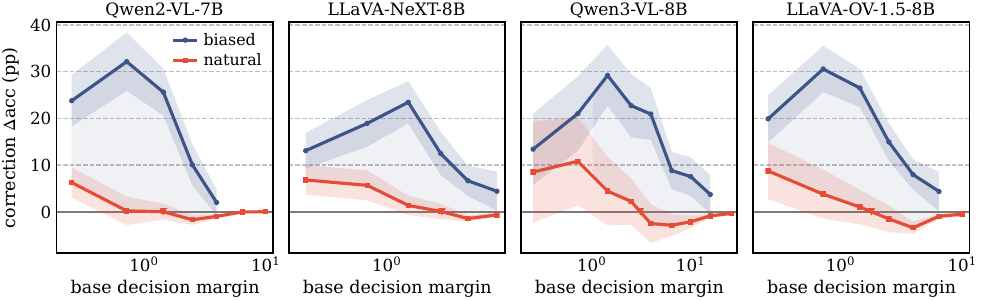}
\caption{Utility of the ungated debiasing step against the base decision
margin, on biased and natural samples. Bands are $95\%$ intervals and shading
marks the margins over which the two curves separate.}
\label{fig:marginlaw}
\vspace{-6mm}
\end{figure*}

The correction of Eq.~\eqref{eq:corr} is charged to every input, yet a
correction of bounded size cannot change a decision the model already makes by
a wide margin. Below that width, whether it helps depends on whether the
prior really drives the input. Therefore, we propose to allocate the
correction by the decision margin, firing it only where the lead of the
predicted answer is small. When it fires, we rebuild the state from anti-prior
contrasts rather than adding to the base scores. A wide-margin input then
keeps its answer at a single forward pass, while a contested one is re-decided
on a state the prior no longer dominates. Rebuilding is necessary because the base
scores are themselves carried by the prior that the branches measure.

Specifically, we define the decision margin \citep{scheffer2001margin} as the
gap between the top two scores, which is formulated as follows:
\begin{equation}
\margin{\scores}=\mathrm{top}_1(\scores)-\mathrm{top}_2(\scores) .
\label{eq:margin}
\end{equation}
A subtractive correction bounded by
$\|\est\|_\infty\le B$ moves the gap between any two scores by at most $2B$.
Therefore, we have
\begin{equation}
\margin{\sbase}>2B \;\Longrightarrow\; \arg\max\scores=\arg\max\sbase
\label{eq:inert}
\end{equation}
and the correction is inert there. The bound is worst-case but informative,
and Appendix~\ref{app:law} shows that it holds on most of the samples \method\
skips. It applies directly to the additive
restoration of \S\ref{sec:restore}, while the state-replacing step here is not
bounded in this sense and its gate is justified instead by measurement. We define the utility $\Util(m)$ as the mean accuracy change when the debiasing
step is applied to every sample of base margin $m$. Figure~\ref{fig:marginlaw}
plots this curve against the base margin for each backbone, on the biased
subsets and on the natural samples separately. On the biased subsets the
correction keeps helping over a wide range of margins. On the natural
samples it already turns harmful at a small margin, where most contested
samples are not prior-driven and the subtraction mostly injects noise. The same
shape repeats on all four backbones, which indicates that the margin is where
the correction should be allocated.

The gate therefore fires only when $\margin{\sbase}\le\ton$. We then form the
visual contrast $\Delta_q=Z(v,q)-Z(v^{\varnothing},q)$, where
$v^{\varnothing}$ is an all-black image. Appendix~\ref{app:manifold} shows that
the branch decides the same way when the blank is replaced by an external
real image. We read the same contrast
through a second wording $q'$, a rephrased prompt that asks for the answer in
a different form. Then, the debiased state is obtained as follows:
\begin{equation}
\scores \gets \max(\Delta_q,\;\Delta_{q'}),
\label{eq:debias}
\end{equation}
where the elementwise maximum keeps the
stronger anti-prior signal per candidate, so the prior cannot suppress a
candidate through one branch alone.

\subsection{Selective Evidence Restoration for Discriminative
Decision-Making}
\label{sec:restore}

Eq.~\eqref{eq:debias} rebuilds the state from contrasts alone, so it discards
the image's own contribution to $Z(v,q)$ along with the prior, and a sample
whose answer turns on that contribution is left without it. Adding it back is
not free either, since a corrupted view carries artifacts of its own that the
subtraction would import. Therefore, we propose to restore a visual residual
from the perturbed view that leaves the model most confident, and to keep it
only where it makes the decision more definite. A view that leaves the model
confident is less likely to have manufactured evidence of its own, so the
residual it yields
carries what the image contributes rather than what the perturbation invents.
A residual that fails to help is discarded instead of applied.

Specifically, let $\scores$ be the state left by the debiasing step of
\S\ref{sec:gating}. When
$\margin{\scores}\le\ttwo$, we select the source by the drop in maximum
softmax probability $\mathrm{mp}(x)=\max_j\mathrm{softmax}(x)_j$ when the
image is replaced by a candidate $v_c$, which is formulated as follows:
\begin{equation}
\Vsrc = \arg\min_{v_c}\ \big[\mathrm{mp}(\sbase)-\mathrm{mp}(Z(v_c,q))\big],
\label{eq:route}
\end{equation}
where the candidate images $v_c$ are grayscale, strong-blur and center-mask,
which remove color, local detail, and the center of the frame respectively, so
that the selection has genuinely different views to choose between. Since
$\mathrm{mp}(\sbase)$ does not depend on $v_c$, Eq.~\eqref{eq:route} keeps the
view whose prediction distribution stays sharpest. Then, we form the residual
as follows:
\begin{equation}
\Dtwo=Z(v,q)-Z(\Vsrc,q).
\label{eq:resid}
\end{equation}
Both terms carry the same question $q$ and therefore the same prior
contribution, which the subtraction cancels. What remains in $\Dtwo$ is what
the real image adds over the perturbed one. The residual is restored only when it
widens the margin by at least $\tauV$, and we have
\begin{equation}
\scores \gets \scores + \Dtwo
\quad\text{if}\ \ \margin{\scores+\Dtwo}-\margin{\scores}\ge\tauV,
\label{eq:commit}
\end{equation}
and a sample whose residual falls short of $\tauV$ exits with its debiased
state.

\begin{figure}[t]\centering
\includegraphics[width=\columnwidth]{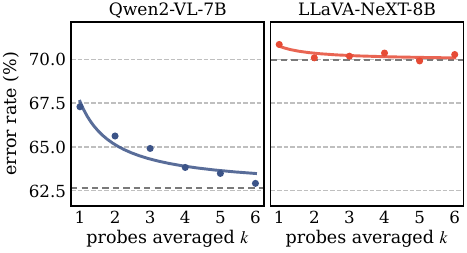}
\caption{Error of the corrector of Eq.~\eqref{eq:corr} as $\est$ averages $k$
branches, with an $a+b/k$ fit and its asymptote $a$ dashed.}
\label{fig:biasvar}
\vspace{-4mm}
\end{figure}

\subsection{Perturbation Branch Pruning with Calibration}
\label{sec:prune}

Both steps draw on perturbation branches, and their number is not a free
parameter to be raised indefinitely. Since $\est$ only approximates the prior,
averaging more branches shrinks the zero-mean component of its error but not
the component the perturbation itself induces, and the return from each
further branch vanishes against its fixed cost. Therefore, we propose to prune
the branch set per model on a calibration set rather than fix it in advance,
keeping a perturbation family only where the prior mass it removes exceeds the
error it injects. Because both the surviving error and the variance above it
are properties of the encoder, this leaves each backbone with the branches it
can still profit from and charges no pass for the ones it cannot.

Formally, write the estimate as $\est=\prior+\noise$, where $\prior$ is the
prior's contribution to the scores and $\noise$ is estimation error, and split
$\noise$ into a zero-mean part with variance $\sigma^2/k$ and a part $b_k$
that averaging does not shrink. The estimate $\est_k$ formed from $k$ branches
then carries the following error:
\begin{equation}
\mathbb{E}\|\est_k-\prior\|^2=\|b_k\|^2+\sigma^2/k .
\label{eq:budget}
\end{equation}
The bias floor $\|b_k\|^2$ survives any budget, and
the $(k{+}1)$-th branch lowers the error by at most
$\sigma^2/(k(k{+}1))$, which vanishes as $k$ grows; see Appendix~\ref{app:budgetproof} for the derivation. Figure~\ref{fig:biasvar} shows both effects on an
observable proxy, the error rate of the corrector of Eq.~\eqref{eq:corr} whose
estimate averages $k$ branches on the biased samples. An $a+b/k$ fit shows the error leveling
off at a large floor rather than falling to zero, with the encoders differing in the
variance above it.

Each candidate perturbation family is therefore
tested on the calibration set, the validation split
of the target distribution, and kept where it
improves the calibration accuracy. The three thresholds are fit on the
same set, $\ttwo$ and $\tauV$ by grid search and $\ton$ as a percentile of the base margin.
\section{Experiments}
\label{sec:experiments}

This section evaluates \method\ against the CD family on biased benchmarks
and on real-world LVLM datasets, across four backbones from two model
generations.

% tab:stats -- DRBench subset sizes (SCI Table-1 replica). Published DRBench
% counts. Overlap = B + S - BS.
\begin{table}[t]
\centering
\setlength{\tabcolsep}{5pt}
\resizebox{\columnwidth}{!}{%
\begin{tabular}{l cccc}
\Xhline{1.2pt}
\headrow & \multicolumn{4}{c}{Subset Size} \\
\cline{2-5}
\headrow & B Subset & S Subset & BS Subset & Overlap \\
\Xhline{1.2pt}
\altrow LLaVA-NeXT (MCQ) & 1810 & 1005 & 2476 & 339 \\
LLaVA-NeXT (Others) & 345 & 582 & 794 & 133 \\
\altrow LLaVA-NeXT (Overall) & 2155 & 1587 & 3270 & 472 \\
\Xhline{0.6pt}
\altrow Qwen2-VL (MCQ) & 1080 & 252 & 1243 & 89 \\
Qwen2-VL (Others) & 327 & 311 & 513 & 125 \\
\altrow Qwen2-VL (Overall) & 1407 & 563 & 1756 & 214 \\
\Xhline{1.2pt}
\end{tabular}}
\caption{The size of each subset in constructed DRBench. The overall
number of samples across all six datasets is $13{,}251$, with the MCQ
and Others categories being $10{,}632$ and $2{,}619$, respectively.}
\label{tab:stats}
\vspace{-4mm}
\end{table}

% tab:main -- DRBench biased subsets, Test split. Baselines are PUBLISHED
% numbers (SCI paper Tables 2/9); MALT = frozen noadm ladder replayed from
% the released bundles. Bold/underline generated from the numbers.
\begin{table*}[t]
\centering
\resizebox{\textwidth}{!}{%
\begin{tabular}{l l ccc c ccc c ccc c c}
\Xhline{1.2pt}
\headrow & & \multicolumn{3}{c}{Bias Subset} && \multicolumn{3}{c}{Sensitivity Subset} && \multicolumn{3}{c}{BS Subset} && \\
\cline{3-5}\cline{7-9}\cline{11-13}
\headrow \multirow{-2}{*}{Model} & \multirow{-2}{*}{Method} & MCQ & Others & All && MCQ & Others & All && MCQ & Others & All && Passes \\
\Xhline{1.2pt}
\altrow  & Base & 5.37 & 8.56 & 6.11 && 38.10 & 34.41 & 36.06 && 10.78 & 23.59 & 14.52 && 1 \\
 & TIE & 16.20 & 16.82 & 16.35 && 45.63 & 36.66 & 40.67 && 20.27 & 27.29 & 22.32 && 2 \\
\altrow  & VCD & 15.74 & 21.71 & 17.13 && 46.83 & 40.84 & 43.52 && 20.11 & 30.41 & 23.12 && 2 \\
 & M3ID & 19.81 & 21.71 & 20.26 && \underline{47.22} & 41.16 & 43.87 && 23.65 & 30.60 & 25.68 && 2 \\
\altrow  & SCI$_3$ & 21.67 & \underline{26.30} & 22.74 && 44.05 & 42.44 & 43.16 && 24.54 & 32.75 & 26.94 && 3 \\
 & SCI$_5$ & 24.91 & 25.69 & 25.09 && \underline{47.22} & 42.44 & 44.58 && 28.00 & 33.14 & 29.50 && 5 \\
\altrow  & SCI$_7$ & \underline{27.04} & \textbf{29.66} & \underline{27.65} && \underline{47.22} & \textbf{45.98} & \underline{46.54} && \underline{29.61} & \textbf{36.84} & \underline{31.72} && 7 \\
\multirow{-8}{*}{Qwen2-VL-7B} & \method & \textbf{30.26} & 25.15 & \textbf{29.07} && \textbf{48.73} & \underline{45.63} & \textbf{47.09} && \textbf{32.30} & \underline{35.16} & \textbf{33.13} && 4.9 \\
\Xhline{1.2pt}
\altrow  & Base & 0.00 & 0.00 & 0.00 && 39.20 & 37.63 & 38.63 && 15.91 & 27.58 & 18.75 && 1 \\
 & TIE & 12.98 & 23.48 & 14.66 && 39.00 & 57.56 & 45.81 && 21.89 & 44.21 & 27.31 && 2 \\
\altrow  & VCD & 12.65 & 25.51 & 14.71 && 40.50 & 56.53 & 46.38 && 22.54 & 44.58 & 27.89 && 2 \\
 & M3ID & 16.91 & 25.22 & 18.24 && 39.90 & 56.36 & 45.94 && 24.15 & 44.33 & 29.05 && 2 \\
\altrow  & SCI$_3$ & 21.22 & 35.36 & 23.48 && 39.60 & \underline{60.31} & 47.20 && 27.14 & 50.13 & 32.72 && 3 \\
 & SCI$_5$ & 23.81 & 37.97 & 26.08 && \underline{40.60} & \textbf{60.65} & \textbf{47.95} && 28.80 & 51.01 & 34.19 && 5 \\
\altrow  & SCI$_7$ & \underline{24.86} & \underline{38.26} & \underline{27.01} && 40.10 & \textbf{60.65} & \underline{47.64} && \underline{29.68} & \underline{51.26} & \underline{34.92} && 7 \\
\multirow{-8}{*}{LLaVA-NeXT-8B} & \method & \textbf{26.95} & \textbf{41.83} & \textbf{29.37} && \textbf{40.99} & 58.30 & 47.29 && \textbf{31.13} & \textbf{52.01} & \textbf{36.22} && 4.7 \\
\Xhline{1.2pt}
\end{tabular}}
\caption{Experiments on the B(ias) Subset, S(ensitivity) Subset, and BS
Subset of DRBench. Baseline results are from \citet{tang2026sci}.
Passes is the mean per-sample forward count. We bold the
best result in each column and underline the second best.}
\label{tab:main}
\end{table*}

% tab:natural -- SCI Table-4 layout: single Method column, Single Dataset /
% Question Type column groups. Base, TIE, VCD, M3ID, SCI5 are SCI's published
% numbers; MALT is ours. Arrows mark the change over the per-backbone base.
% Shading (headrow/altrow) matches tab:main.
\begin{table*}[t]
\centering
\resizebox{\textwidth}{!}{%
\begin{tabular}{l cccccc ccc}
\Xhline{1.2pt}
\headrow & \multicolumn{6}{c}{Single Dataset} & \multicolumn{3}{c}{Question Type} \\
\cline{2-7}\cline{8-10}
\headrow \multirow{-2}{*}{Method} & MMB-C & MMB-E & MME & CCB & MMS & ViLP & MCQ & Others & Overall \\
\Xhline{1.2pt}
\altrow Qwen2-VL & 85.26 & 86.36 & 87.89 & 73.22 & 59.50 & 56.53 & 80.91 & 79.27 & 80.58 \\
Qwen2-VL-TIE & 86.00\up{0.74} & 86.59\up{0.23} & 86.52\down{1.37} & 73.84\up{0.62} & 59.00\down{0.50} & 57.08\up{0.55} & 81.30\up{0.39} & 78.43\down{0.84} & 80.73\up{0.15} \\
\altrow Qwen2-VL-VCD & 86.05\up{0.79} & 86.56\up{0.20} & 86.41\down{1.48} & 73.77\up{0.55} & 60.08\up{0.58} & 57.92\up{1.39} & 81.42\up{0.51} & 78.58\down{0.69} & 80.86\up{0.28} \\
Qwen2-VL-M3ID & 85.69\up{0.43} & 86.46\up{0.10} & 86.10\down{1.79} & 73.96\up{0.74} & 59.75\up{0.25} & 57.78\up{1.25} & 81.25\up{0.34} & 78.31\down{0.96} & 80.67\up{0.09} \\
\altrow Qwen2-VL-SCI$_5$ & 85.97\up{0.71} & 86.67\up{0.31} & 87.36\down{0.53} & 73.59\up{0.37} & 59.92\up{0.42} & 58.06\up{1.53} & 81.39\up{0.48} & 79.31\up{0.04} & 80.98\up{0.40} \\
Qwen2-VL-\method\,(ours) & 86.05\up{0.79} & 86.49\up{0.13} & 88.26\up{0.37} & 73.90\up{0.68} & 60.68\up{1.18} & 56.53 & 81.49\up{0.58} & 79.53\up{0.26} & 81.10\up{0.52} \\
\Xhline{0.6pt}
\altrow LLaVA-NeXT & 78.00 & 79.72 & 79.57 & 47.00 & 44.75 & 51.53 & 70.12 & 71.86 & 70.46 \\
LLaVA-NeXT-TIE & 78.28\up{0.28} & 80.28\up{0.56} & 77.30\down{2.27} & 45.65\down{1.35} & 46.00\up{1.25} & 53.19\up{1.66} & 70.36\up{0.24} & 70.68\down{1.18} & 70.42\down{0.04} \\
\altrow LLaVA-NeXT-VCD & 78.38\up{0.38} & 80.28\up{0.56} & 78.09\down{1.48} & 46.63\down{0.37} & 45.00\up{0.25} & 54.31\up{2.78} & 70.44\up{0.32} & 71.55\down{0.31} & 70.66\up{0.20} \\
LLaVA-NeXT-M3ID & 78.31\up{0.31} & 80.18\up{0.46} & 78.62\down{0.95} & 45.89\down{1.11} & 45.92\up{1.17} & 54.03\up{2.50} & 70.36\up{0.24} & 71.86 & 70.66\up{0.20} \\
\altrow LLaVA-NeXT-SCI$_5$ & 78.21\up{0.21} & 80.08\up{0.36} & 80.15\up{0.58} & 46.20\down{0.80} & 45.75\up{1.00} & 53.06\up{1.53} & 70.32\up{0.20} & 72.70\up{0.84} & 70.79\up{0.33} \\
LLaVA-NeXT-\method\,(ours) & 78.21\up{0.21} & 80.26\up{0.54} & 80.62\up{1.05} & 47.79\up{0.79} & 46.74\up{1.99} & 52.78\up{1.25} & 70.74\up{0.62} & 72.97\up{1.11} & 71.18\up{0.72} \\
\Xhline{1.2pt}
\end{tabular}}
\caption{Experiments on the six real-world LVLM datasets, MMB(ench-Dev)-C/E(N-V11),
MME, CCB(ench), MMS(tar), and ViLP (using 80\% test splits). Baseline results are from \citet{tang2026sci}.}
\label{tab:natural}
\vspace{-4mm}
\end{table*}

\subsection{Experimental Settings}

\noindent\textbf{Datasets and Metrics.} We evaluate on DRBench
\citep{tang2026sci}, whose biased subsets are built from MMStar
\citep{chen2024mmstar}, CCBench and MMBench-EN/CN \citep{liu2024mmbench}, MME
\citep{fu2023mme}, and ViLP \citep{luo2024vilp}. The evaluation subsets are
B(ias), where the model answers wrongly and repeats that answer under a blank
image, S(ensitivity), where its answer changes across prompt rephrasings, and
their union BS. They are built with visual and textual counterfactual probes (VCF
and TCF), with MCQ, Others, and Overall accuracy on the BS subset, following
the benchmark. We report top-1 accuracy throughout,
scoring open-ended items by generation with prefix match. MME uses its own
score, which we convert to accuracy so that it combines with the other
datasets.
Besides accuracy, we report each method's average number of forward passes per
sample as its cost. Table~\ref{tab:stats} summarizes the subset statistics.

\noindent\textbf{Backbones.} We use the Hugging Face releases of Qwen2-VL-7B
\citep{wang2024qwen2vl} and LLaVA-NeXT-8B \citep{liu2024llavanext} as base
models, and extend the evaluation to Qwen3-VL-8B \citep{bai2025qwen3vl} and
LLaVA-OneVision-1.5-8B \citep{an2025llavaov15}. Qwen2-VL follows its release default of bfloat16 precision with
top-$k$ decoding and LLaVA-NeXT its default of float16 with greedy decoding,
and the two newer backbones run in bfloat16 with greedy decoding. 

\noindent\textbf{Counterfactual Construction.} Every branch is built from the
same question and image. The blank branch replaces the image with an all-black
canvas of the same size. The visual sources of Eq.~\eqref{eq:route} transform
the real image rather than discarding it, by conversion to grayscale, by a
Gaussian blur of radius $100$, and by covering the central half of the frame
with the image's mean color. The answer-format branch keeps the image and
swaps the instruction for a second wording of the same answer-format
requirement, so contrasting it against the same wording under a blank image
measures the prior through a different phrasing.

\noindent\textbf{Baselines.} The CD baselines instantiate
the prior-estimate subtraction of Eq.~\eqref{eq:corr} with different
counterfactual sets. TIE removes one visual counterfactual, VCD and M3ID
reweight that same difference with M3ID growing the weight along the token
position, and SCI$_{3/5/7}$ average $M{=}N{=}1/2/3$ visual and textual
counterfactuals together with the original input. Their branches follow their
own papers. SCI uses a blank and a noised image as its visual counterfactuals
and two prompt rephrasings as its textual ones. On the two newer backbones
we run every baseline with the implementation released with DRBench, validated
against the released cells, and score open-ended items over the shared
candidate pool.

\begin{figure}[t]\centering
\includegraphics[width=\columnwidth]{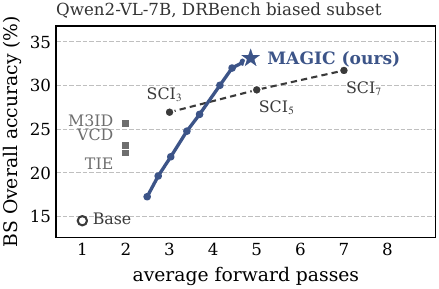}
\caption{Accuracy against compute on the biased subset of Qwen2-VL as $t_1$
is swept. Each baseline sits at one fixed budget, while \method\ traces a
frontier. The star marks the deployed setting.}
\label{fig:frontier}
\vspace{-4mm}
\end{figure}

\subsection{Performance Comparison}
\noindent\textbf{Results on DRBench.} We compare the \method\ with
the CD family on the B(ias), S(ensitivity), and BS subsets of DRBench across
two backbones, and report the accuracy together with the average number of
forward passes per sample in Table~\ref{tab:main}. According to the results, we
make several observations (Obs.) listed as follows:
\textbf{Obs.\ding{182} \method\ attains the best BS Overall on both backbones,
outperforming the uncorrected base model and the strongest two-branch
correctors.} As shown in Table~\ref{tab:main}, it improves over the base model in terms of BS Overall accuracy
by $18.61\%$ on Qwen2-VL and $17.47\%$  on
LLaVA-NeXT, and it surpasses M3ID by $7.45\%$ and $7.17\%$.
\textbf{Obs.\ding{183} \method\ improves over the SCI family at a lower
inference cost.} The results in Table~\ref{tab:main} demonstrate that \method\
improves over the cost-matched $\text{SCI}_5$ by $3.63\%$ and $2.03\%$, and
exceeds the seven-branch $\text{SCI}_7$ on Overall while using roughly $70\%$
of its passes.
We further sweep the debiasing gate $\ton$ on the biased subset of Qwen2-VL and
plot accuracy against the average number of forward passes in
Figure~\ref{fig:frontier}. According to the results, each baseline sits at one
fixed budget, while \method\ traces the accuracy--compute frontier over $\ton$.

\noindent\textbf{Results on Real-world LVLM Datasets.} We evaluate \method\
against the CD family on the six real-world LVLM datasets across two
backbones, and report the accuracy. As shown in
Table~\ref{tab:natural}, \method\ improves over the base on both backbones
and outperforms the strongest CD baseline's Overall at fewer passes, while
TIE and VCD lose accuracy on the Others question type. The margins are small
because vulnerable samples are only a fraction of these datasets.

\begin{figure*}[!t]\centering
\begin{minipage}[t]{0.485\textwidth}\centering
  {\small\textbf{(a) One gate opens \,---\, $4$ forward passes}}\par\vspace{2pt}
  \casehdr{}
    {Which is the main topic of the image?}
    {A. A driving car & B. Driving cars \\ C. Driving buses & D. A driving bus}
    {C. Driving buses}
  \vspace{2pt}
  \casetab{%
   \altrow Base $\sbase$ & $-5.04$ & $-5.29$ & $-0.91$ & $\mathbf{-0.54}$ & $0.37$ \\
   Debiased $\Done$ & $-3.09$ & $-3.71$ & $\mathbf{2.54}$ & $0.37$ & $2.17$ \\
   \altrow Residual $\Dtwo$ & \multicolumn{4}{c}{gate closed} & \\
   Final $\scores$ & $-3.09$ & $-3.71$ & $\mathbf{2.54}$ & $0.37$ & $2.17$ \\}
  \casenote{stgAfill}{stgAline}{\textbf{Base $\to$ D}~\xmark\quad Several
    buses share the street and the model answers as if there were one, $0.37$
    ahead of the plural option. The margin $0.37\le\ton{=}3.25$ opens the
    debiasing gate.}
  \casenote{stgBfill}{stgBline}{\textbf{Step 1 $\to$ C}~\okmark\quad
    Contrasting the image against a blank one under both instructions removes
    what the question alone makes likely. C gains $3.45$ against D's $0.91$ and
    takes the lead, at a margin of $2.17$.}
  \casenoteoff{stgCfill}{stgCline}{\textbf{Step 2: not run}\quad That
    $2.17>\ttwo{=}0.4$, so the restoration gate stays shut and no visual
    residual is built. The sample leaves after $4$ passes, three of them spent
    by the step that fired.}
\end{minipage}
\hfill
\begin{minipage}[t]{0.485\textwidth}\centering
  {\small\textbf{(b) Both gates open \,---\, $7$ forward passes}}\par\vspace{2pt}
  \casehdr{}
    {Which image is the brightest one?}
    {A. upper left & B. upper right \\ C. down left & D. down right}
    {C. down left}
  \vspace{2pt}
  \casetab{%
   \altrow Base $\sbase$ & $\mathbf{-0.22}$ & $-5.22$ & $-1.72$ & $-4.59$ & $1.50$ \\
   Debiased $\Done$ & $\mathbf{0.40}$ & $-3.22$ & $0.35$ & $-2.15$ & $0.05$ \\
   \altrow Residual $\Dtwo$ & $-0.10$ & $-0.22$ & $0.65$ & $-0.22$ & \\
   Final $\scores$ & $0.30$ & $-3.44$ & $\mathbf{1.00}$ & $-2.37$ & $0.70$ \\}
  \casenote{stgAfill}{stgAline}{\textbf{Base $\to$ A}~\xmark\quad The model
    picks the upper-left frame, a photograph shot in the dark, rather than the
    lit one below it. The margin $1.50\le\ton{=}3.25$ opens the debiasing
    gate.}
  \casenote{stgBfill}{stgBline}{\textbf{Step 1 $\to$ A}~\xmark\quad
    Debiasing lifts C by $2.07$ but leaves it $0.05$ behind A, so the answer
    does not settle. That $0.05\le\ttwo{=}0.4$ opens the restoration gate.}
  \casenote{stgCfill}{stgCline}{\textbf{Step 2 $\to$ C}~\okmark\quad The
    residual $\sbase-\Vsrc$ puts the removed evidence back: it adds $0.65$ to C
    and moves the rest down. The margin widens by $0.65\ge\tauV{=}0.625$, so
    the step commits.}
\end{minipage}
\caption{Two biased-subset samples \method\ repairs on Qwen2-VL, at the fitted
thresholds. Each note keeps its stage's color from Figure~1 and
is dashed when that gate never opened. \okmark\ and \xmark\ flag the answer a
stage leaves, $m$ is the top-two margin, and the committed answer is bold.}
\label{fig:casebody}
\vspace{-4mm}
\end{figure*}

\noindent\textbf{Results on Newer Backbones.} We extend the evaluation to two
more recent models. As shown in Table~\ref{tab:newgen}, \method\ attains the
best B and BS on both and the best S on Qwen3-VL. It improves BS Overall over
the base by $19.28\%$ on Qwen3-VL and $15.58\%$ on LLaVA-OneVision-1.5,
over M3ID by $16.66\%$ and $7.81\%$, and over the best SCI setting by $5.40\%$ and
$0.92\%$. On both backbones the calibration screen keeps a single debiasing contrast, and on
Qwen3-VL it also drops the restoration step, so the correction costs $2.9$
and $4.0$ passes. The SCI family itself stops scaling here, with
$\text{SCI}_7$ falling below $\text{SCI}_5$ on Qwen3-VL and below
$\text{SCI}_3$ on LLaVA-OneVision-1.5 in BS Overall, the vanishing return
that Eq.~\eqref{eq:budget} predicts.

\begin{table}[t]
\centering
\renewcommand{\arraystretch}{1.05}
\resizebox{\columnwidth}{!}{%
\begin{tabular}{l l cccc}
\Xhline{1.2pt}
\headrow Model & Method & B & S & BS & Passes \\
\Xhline{1.2pt}
\altrow & Base & 0.49 & 34.65 & 12.56 & 1 \\
 & TIE & 5.71 & 32.44 & 13.47 & 2 \\
\altrow & VCD & 7.19 & 32.60 & 14.27 & 2 \\
 & M3ID & 8.74 & 33.54 & 15.18 & 2 \\
\altrow & SCI$_3$ & 21.37 & 41.42 & 24.96 & 3 \\
 & SCI$_5$ & \underline{22.43} & \underline{47.56} & \underline{26.44} & 5 \\
\altrow & SCI$_7$ & 21.93 & 46.61 & 25.92 & 7 \\
 \multirow{-8}{*}{Qwen3-VL} & \method & \textbf{28.98} & \textbf{49.13} & \textbf{31.84} & 2.9 \\
\Xhline{1.2pt}
\altrow & Base & 0.17 & 52.16 & 26.58 & 1 \\
 & TIE & 11.75 & 52.81 & 31.60 & 2 \\
\altrow & VCD & 12.03 & 53.09 & 32.00 & 2 \\
 & M3ID & 15.96 & 53.60 & 34.35 & 2 \\
\altrow & SCI$_3$ & 25.24 & \textbf{57.99} & \underline{41.24} & 3 \\
 & SCI$_5$ & 24.85 & 52.73 & 38.82 & 5 \\
\altrow & SCI$_7$ & \underline{25.74} & 52.23 & 38.82 & 7 \\
 \multirow{-8}{*}{OneVision-1.5} & \method & \textbf{27.49} & \underline{56.47} & \textbf{42.16} & 4.0 \\
\Xhline{1.2pt}
\end{tabular}}
\caption{B/S/BS Overall accuracy on two more recent backbones. We bold the best result in
each column and underline the second best.}
\label{tab:newgen}
\vspace{-4mm}
\end{table}

\subsection{Further Analysis}

\noindent\textbf{Ablation Study.} We then investigate how the two gated steps and the two debiasing contrasts in
\method\ affect the final performance. In particular, we design four variants,
which remove the debiasing step, the restoration step, the blank contrast, and
the answer-format contrast, respectively. We perform experiments on the BS
subset of DRBench with both backbones, and the results are shown in
Table~\ref{tab:ablation}, from which we have the following observations:
\textbf{Obs.\ding{184} Both gated steps contribute to the final performance.}
Removing the debiasing step collapses the BS accuracy, since the visual
residual is left with no cleaned state to build on, and adding the residual
back on top of it recovers accuracy on the multiple-choice items.
\textbf{Obs.\ding{185} The two debiasing contrasts are complementary.} Within
the debiasing step, replacing the fusion of the blank and the answer-format
contrasts by either one alone costs between $0.5\%$ and $1.7\%$ of BS Overall,
so each step carries a distinct part of the result.

% tab:ablation -- pure leave-one-out on both backbones, BS subset
% (MCQ/Others/Overall) plus the mean per-sample forward count. Accuracy
% replayed from the frozen bundles (malt.py); the Qwen rows and both Overall
% columns reproduce the published cells. Passes from
% research/malt_ablation_passes.py. Shading matches tab:main (gray/white/gray).
\begin{table}[t]
\centering
\renewcommand{\arraystretch}{1.15}
\setlength{\tabcolsep}{5pt}
\resizebox{\columnwidth}{!}{%
\begin{tabular}{l l cccc}
\Xhline{1.2pt}
\headrow Model & Configuration & MCQ & Others & Overall & Passes \\
\Xhline{1.2pt}
\altrow & \method\ w/o CD & 13.17 & 25.39 & 16.70 & 2.2 \\
 & \method\ w/o BC & 31.59 & 34.77 & 32.51 & 4.0 \\
\altrow & \method\ w/o AF & 31.11 & 32.42 & 31.49 & 3.2 \\
 & \method\ w/o SR & 30.40 & 33.59 & 31.32 & 3.8 \\
\altrow \multirow{-5}{*}{Qwen2-VL} & \method & 32.30 & 35.16 & 33.13 & 4.9 \\
\Xhline{0.6pt}
\altrow & \method\ w/o CD & 17.07 & 38.44 & 22.28 & 1.9 \\
 & \method\ w/o BC & 30.48 & 51.88 & 35.70 & 3.7 \\
\altrow & \method\ w/o AF & 30.44 & 50.88 & 35.43 & 3.0 \\
 & \method\ w/o SR & 30.52 & 52.76 & 35.95 & 3.9 \\
\altrow \multirow{-5}{*}{LLaVA-NeXT} & \method & 31.13 & 52.01 & 36.22 & 4.7 \\
\Xhline{1.2pt}
\end{tabular}}
\caption{Ablation on the BS Subset of DRBench. CD: Contrastive Debiasing, BC: Blank
Contrast, AF: Answer-Format Contrast, SR: Selective Restoration. Passes is the
mean per-sample forward count.}
\label{tab:ablation}
\vspace{-5mm}
\end{table}

\begin{table}[!t]
\centering
\renewcommand{\arraystretch}{1.0}
\setlength{\tabcolsep}{5pt}
\resizebox{\columnwidth}{!}{%
\begin{tabular}{l c cccc}
\Xhline{1.2pt}
\headrow Model & Value & MCQ & Others & Overall & Passes \\
\Xhline{1.2pt}
\multicolumn{6}{l}{\emph{Debiasing gate $t_1$}} \\
\altrow & 0.5 & 18.81 & 29.30 & 21.84 & 3.0 \\
 & 1.0 & 24.68 & 31.64 & 26.69 & 3.7 \\
\altrow & 2.0 & 31.03 & 34.38 & 32.00 & 4.4 \\
 & 3.0 & 32.30 & 34.96 & 33.07 & 4.8 \\
\altrow & 5.0 & 32.38 & 35.74 & 33.35 & 5.0 \\
 \multirow{-6}{*}{Qwen2-VL} & 8.0 & \textbf{32.46} & \textbf{35.94} & \textbf{33.47} & 5.0 \\
\Xhline{1.2pt}
\multicolumn{6}{l}{\emph{Restoration gate $t_2$}} \\
\altrow & 0.0 & 30.32 & 33.59 & 31.26 & 4.0 \\
 & 0.2 & 31.35 & 33.40 & 31.94 & 4.4 \\
\altrow & 0.3 & 32.14 & 33.59 & 32.56 & 4.6 \\
 & 0.4 & 32.30 & 34.18 & 32.84 & 4.9 \\
\altrow & 0.6 & \textbf{32.46} & \textbf{35.16} & \textbf{33.24} & 5.1 \\
 \multirow{-6}{*}{Qwen2-VL} & 1.0 & 32.06 & \textbf{35.16} & 32.96 & 5.5 \\
\Xhline{1.2pt}
\end{tabular}}
\caption{Sensitivity to the two allocation thresholds on Qwen2-VL, BS Subset
of DRBench. We bold the best result in each column within a block.}
\label{tab:sensitivity}
\vspace{-4mm}
\end{table}

\noindent\textbf{Sensitivity Analysis.} To see how much the result
depends on the fitted constants, we sweep the two
allocation thresholds one at a time on Qwen2-VL
in Table~\ref{tab:sensitivity}. Above the fitted value $\ton{=}3.25$, accuracy
is flat, gaining at most $0.4\%$ in BS Overall as $\ton$ rises to $8$,
and it falls off only below it, where the gate stops
firing on samples that need it. That plateau is the
inert regime of Eq.~\eqref{eq:inert}, where a larger $\ton$ mostly
admits samples too confident to move. In deployment $\ton$ is not tuned for accuracy but set as the
$95$th percentile of the base margin on the biased
calibration set. The restoration gate $\ttwo$ moves BS
Overall by $1.98\%$ on Qwen2-VL while the mean
pass count rises from $4.0$ to $5.5$, so it trades additional compute for accuracy within a narrow band.

\noindent\textbf{Case Study.} To see how the allocation rule behaves on individual
samples, we trace two Qwen2-VL samples through \method.
Figure~\ref{fig:casebody} presents two cases the base model answers incorrectly. In the first the image shows several buses and the
model answers as if there were one, $0.37$ ahead of the correct option. The
debiasing step fires and moves the answer to that option with a margin of
$2.17$, wide enough that the restoration step never runs. In the second the
debiasing step leaves the top two options $0.05$ apart, so the restoration step
also fires and the visual residual adds $0.65$ to the correct option. The same
rule therefore spends four forward passes on one sample and seven on the
other. A fixed branch set would instead spend the same number on both,
including on the first sample, where one step already leaves the margin wide. 
\section{Conclusion}
\label{sec:conclusion}

In this paper, we propose \method, a training-free corrector for test-time
robustness scaling that fires only where the decision margin is small,
restores the visual evidence the perturbation removes, and sizes its branch
set per model. Two measurements on the contrastive decoding family motivate
this design. Extensive experiments demonstrate the effectiveness of the proposed \method{}.

\section*{Limitations}
\label{sec:limits}
% Limitations (unnumbered; \section*{Limitations} in main.tex).
\method\ operates on the model's scores over a fixed candidate set, the
option letters for multiple-choice items and a materialized pool of
generations for open-ended ones, and does not intervene inside
token-by-token decoding, so free-form captioning benchmarks
fall outside the scope of these results. Its three thresholds are fitted per target distribution on a validation split, so a new distribution requires re-running that selection.
Extending margin allocation into token-by-token decoding, so that it
also covers free-form generation, is the natural next step for this
line of work.

\bibliography{references}

@article{tang2026sci,
  title={Scaling Test-Time Robustness of Vision-Language Models via Self-Critical Inference Framework},
  author={Tang, Kaihua and Qi, Jiaxin and Ou, Jinli and Zheng, Yuhua and Huang, Jianqiang},
  journal={arXiv preprint arXiv:2603.07659},
  year={2026}
}

@inproceedings{leng2024vcd,
  title={Mitigating object hallucinations in large vision-language models through visual contrastive decoding},
  author={Leng, Sicong and Zhang, Hang and Chen, Guanzheng and Li, Xin and Lu, Shijian and Miao, Chunyan and Bing, Lidong},
  booktitle={2024 IEEE/CVF Conference on Computer Vision and Pattern Recognition (CVPR)},
  pages={13872--13882},
  year={2024},
  organization={IEEE}
}

@inproceedings{favero2024m3id,
  title={Multi-modal hallucination control by visual information grounding},
  author={Favero, Alessandro and Zancato, Luca and Trager, Matthew and Choudhary, Siddharth and Perera, Pramuditha and Achille, Alessandro and Swaminathan, Ashwin and Soatto, Stefano},
  booktitle={2024 IEEE/CVF Conference on Computer Vision and Pattern Recognition (CVPR)},
  pages={14303--14312},
  year={2024},
  organization={IEEE}
}

@inproceedings{niu2021counterfactual,
  title={Counterfactual vqa: A cause-effect look at language bias},
  author={Niu, Yulei and Tang, Kaihua and Zhang, Hanwang and Lu, Zhiwu and Hua, Xian-Sheng and Wen, Ji-Rong},
  booktitle={2021 IEEE/CVF Conference on Computer Vision and Pattern Recognition (CVPR)},
  pages={12695--12705},
  year={2021},
  organization={IEEE}
}

@inproceedings{tang2020unbiased,
  title={Unbiased scene graph generation from biased training},
  author={Tang, Kaihua and Niu, Yulei and Huang, Jianqiang and Shi, Jiaxin and Zhang, Hanwang},
  booktitle={2020 IEEE/CVF Conference on Computer Vision and Pattern Recognition (CVPR)},
  pages={3713--3722},
  year={2020},
  organization={IEEE}
}

@article{huang2026chasd,
  title={CHASD: Language Increment-Calibrated Contrastive Decoding against Hallucination in LVLMs},
  author={Huang, Xiaoyi and Zhang, Kejia and Luo, Zhiming},
  journal={arXiv preprint arXiv:2605.23344},
  year={2026}
}

@article{liu2026htdc,
  title={HTDC: Hesitation-Triggered Differential Calibration for Mitigating Hallucination in Large Vision-Language Models},
  author={Liu, Xinyun},
  journal={arXiv preprint arXiv:2604.12115},
  year={2026}
}

@article{ren2026nolan,
  title={NoLan: Mitigating Object Hallucinations in Large Vision-Language Models via Dynamic Suppression of Language Priors},
  author={Ren, Lingfeng and Yu, Weihao and Yu, Runpeng and Wang, Xinchao},
  journal={arXiv preprint arXiv:2602.22144},
  year={2026}
}

@inproceedings{antol2015vqa,
  title={Vqa: Visual question answering},
  author={Antol, Stanislaw and Agrawal, Aishwarya and Lu, Jiasen and Mitchell, Margaret and Batra, Dhruv and Lawrence Zitnick, C and Parikh, Devi},
  booktitle={Proceedings of the IEEE international conference on computer vision},
  pages={2425--2433},
  year={2015}
}

@article{goyal2017making,
  title={Making the v in vqa matter: Elevating the role of image understanding in visual question answering},
  author={Goyal, Yash and Khot, Tejas and Agrawal, Aishwarya and Summers-Stay, Douglas and Batra, Dhruv and Parikh, Devi},
  journal={International Journal of Computer Vision},
  volume={127},
  number={4},
  pages={398--414},
  year={2019},
  publisher={Springer}
}

@inproceedings{agrawal2018dont,
  title={Don't just assume; look and answer: Overcoming priors for visual question answering},
  author={Agrawal, Aishwarya and Batra, Dhruv and Parikh, Devi and Kembhavi, Aniruddha},
  booktitle={2018 IEEE/CVF Conference on Computer Vision and Pattern Recognition},
  pages={4971--4980},
  year={2018},
  organization={IEEE}
}

@article{fu2023mme,
  title={Mme: A comprehensive evaluation benchmark for multimodal large language models},
  author={Fu, Chaoyou and Chen, Peixian and Shen, Yunhang and Qin, Yulei and Zhang, Mengdan and Lin, Xu and Yang, Jinrui and Zheng, Xiawu and Li, Ke and Sun, Xing and others},
  journal={Advances in Neural Information Processing Systems},
  volume={38},
  year={2026}
}

@article{chen2024mmstar,
  title={Are we on the right way for evaluating large vision-language models?},
  author={Chen, Lin and Li, Jinsong and Dong, Xiaoyi and Zhang, Pan and Zang, Yuhang and Chen, Zehui and Duan, Haodong and Wang, Jiaqi and Qiao, Yu and Lin, Dahua and others},
  journal={Advances in Neural Information Processing Systems},
  volume={37},
  pages={27056--27087},
  year={2024}
}

@inproceedings{liu2024mmbench,
  title={Mmbench: Is your multi-modal model an all-around player?},
  author={Liu, Yuan and Duan, Haodong and Zhang, Yuanhan and Li, Bo and Zhang, Songyang and Zhao, Wangbo and Yuan, Yike and Wang, Jiaqi and He, Conghui and Liu, Ziwei and others},
  booktitle={European conference on computer vision},
  pages={216--233},
  year={2024},
  organization={Springer}
}

@article{luo2024vilp,
  title={Probing visual language priors in vlms},
  author={Luo, Tiange and Cao, Ang and Lee, Gunhee and Johnson, Justin and Lee, Honglak},
  journal={arXiv preprint arXiv:2501.00569},
  year={2024}
}

@article{wang2024qwen2vl,
  title={Qwen2-vl: Enhancing vision-language model's perception of the world at any resolution},
  author={Wang, Peng and Bai, Shuai and Tan, Sinan and Wang, Shijie and Fan, Zhihao and Bai, Jinze and Chen, Keqin and Liu, Xuejing and Wang, Jialin and Ge, Wenbin and others},
  journal={arXiv preprint arXiv:2409.12191},
  year={2024}
}

@misc{liu2024llavanext,
  title={Llavanext: Improved reasoning, ocr, and world knowledge},
  author={Liu, Haotian and Li, Chunyuan and Li, Yuheng and Li, Bo and Zhang, Yuanhan and Shen, Sheng and Lee, Yong Jae},
  year={2024}
}

@article{bai2025qwen3vl,
  title={Qwen3-vl technical report},
  author={Bai, Shuai and Cai, Yuxuan and Chen, Ruizhe and Chen, Keqin and Chen, Xionghui and Cheng, Zesen and Deng, Lianghao and Ding, Wei and Gao, Chang and Ge, Chunjiang and others},
  journal={arXiv preprint arXiv:2511.21631},
  year={2025}
}

@article{an2025llavaov15,
  title={Llava-onevision-1.5: Fully open framework for democratized multimodal training},
  author={An, Xiang and Xie, Yin and Yang, Kaicheng and Zhang, Wenkang and Zhao, Xiuwei and Cheng, Zheng and Wang, Yirui and Xu, Songcen and Chen, Changrui and Zhu, Didi and others},
  journal={arXiv preprint arXiv:2509.23661},
  year={2025}
}

@inproceedings{li2023contrastive,
  title={Contrastive decoding: Open-ended text generation as optimization},
  author={Li, Xiang Lisa and Holtzman, Ari and Fried, Daniel and Liang, Percy and Eisner, Jason and Hashimoto, Tatsunori B and Zettlemoyer, Luke and Lewis, Mike},
  booktitle={Proceedings of the 61st annual meeting of the association for computational linguistics (volume 1: Long papers)},
  pages={12286--12312},
  year={2023}
}

@inproceedings{chuang2024dola,
  title={Dola: Decoding by contrasting layers improves factuality in large language models},
  author={Chuang, Yung-Sung and Xie, Yujia and Luo, Hongyin and Kim, Yoon and Glass, James R and He, Pengcheng},
  booktitle={International Conference on Learning Representations},
  volume={2024},
  pages={54158--54183},
  year={2024}
}

@inproceedings{liu2024pai,
  title={Paying more attention to image: A training-free method for alleviating hallucination in lvlms},
  author={Liu, Shi and Zheng, Kecheng and Chen, Wei},
  booktitle={European Conference on Computer Vision},
  pages={125--140},
  year={2024},
  organization={Springer}
}

@inproceedings{huo2025sid,
  title={Self-introspective decoding: Alleviating hallucinations for large vision-language models},
  author={Huo, Fushuo and Xu, Wenchao and Zhang, Zhong and Wang, Haozhao and Chen, Zhicheng and Zhao, Peilin},
  booktitle={International Conference on Learning Representations},
  volume={2025},
  pages={24272--24295},
  year={2025}
}

@article{deng2025maskcd,
  title={Maskcd: Mitigating lvlm hallucinations by image head masked contrastive decoding},
  author={Deng, Jingyuan and Yang, Yujiu},
  journal={arXiv preprint arXiv:2510.02790},
  year={2025}
}

@article{tang2026ccd,
  title={Thinking by Subtraction: Confidence-Driven Contrastive Decoding for LLM Reasoning},
  author={Tang, Lexiang and Gao, Weihao and Zhao, Bingchen and Ma, Lu and Yang, Bang and Zou, Yuexian and others},
  journal={arXiv preprint arXiv:2602.18232},
  year={2026}
}

@article{qin2026sira,
  title={Do We Really Need External Tools to Mitigate Hallucinations? SIRA: Shared-Prefix Internal Reconstruction of Attribution},
  author={Qin, Tian and Chen, Junzhe and Shi, Yuqing and Zhang, Tianshu and Ju, Qiang and Wen, Lijie},
  journal={arXiv preprint arXiv:2605.14621},
  year={2026}
}

@article{zhao2025cicd,
  title={Cross-image contrastive decoding: Precise, lossless suppression of language priors in large vision-language models},
  author={Zhao, Jianfei and Zhang, Feng and Sun, Xin and Kong, Lingxing and Tan, Zhixing and Feng, Chong},
  journal={arXiv preprint arXiv:2505.10634},
  year={2025}
}

@inproceedings{scheffer2001margin,
  title={Active hidden markov models for information extraction},
  author={Scheffer, Tobias and Decomain, Christian and Wrobel, Stefan},
  booktitle={International symposium on intelligent data analysis},
  pages={309--318},
  year={2001},
  organization={Springer}
}

@inproceedings{utama2020mind,
  title={Mind the trade-off: Debiasing NLU models without degrading the in-distribution performance},
  author={Utama, Prasetya Ajie and Moosavi, Nafise Sadat and Gurevych, Iryna},
  booktitle={Proceedings of the 58th Annual Meeting of the Association for Computational Linguistics},
  pages={8717--8729},
  year={2020}
}

@inproceedings{clark2019dont,
  title={Don’t take the easy way out: Ensemble based methods for avoiding known dataset biases},
  author={Clark, Christopher and Yatskar, Mark and Zettlemoyer, Luke},
  booktitle={Proceedings of the 2019 conference on empirical methods in natural language processing and the 9th international joint conference on natural language Processing (EMNLP-IJCNLP)},
  pages={4069--4082},
  year={2019}
}

@article{cadene2019rubi,
  title={Rubi: Reducing unimodal biases for visual question answering},
  author={Cadene, Remi and Dancette, Corentin and Cord, Matthieu and Parikh, Devi and others},
  journal={Advances in neural information processing systems},
  volume={32},
  year={2019}
}

@inproceedings{holtzman2021surface,
  title={Surface form competition: Why the highest probability answer isn’t always right},
  author={Holtzman, Ari and West, Peter and Shwartz, Vered and Choi, Yejin and Zettlemoyer, Luke},
  booktitle={Proceedings of the 2021 conference on empirical methods in natural language processing},
  pages={7038--7051},
  year={2021}
}

@inproceedings{zhao2021calibrate,
  title={Calibrate before use: Improving few-shot performance of language models},
  author={Zhao, Zihao and Wallace, Eric and Feng, Shi and Klein, Dan and Singh, Sameer},
  booktitle={International conference on machine learning},
  pages={12697--12706},
  year={2021},
  organization={Pmlr}
}

@article{schick2021selfdiag,
  title={Self-diagnosis and self-debiasing: A proposal for reducing corpus-based bias in nlp},
  author={Schick, Timo and Udupa, Sahana and Sch{\"u}tze, Hinrich},
  journal={Transactions of the Association for Computational Linguistics},
  volume={9},
  pages={1408--1424},
  year={2021},
  publisher={MIT Press One Rogers Street, Cambridge, MA 02142-1209, USA journals-info~…}
}

@article{hendrycks2017msp,
  title={A baseline for detecting misclassified and out-of-distribution examples in neural networks},
  author={Hendrycks, Dan and Gimpel, Kevin},
  journal={arXiv preprint arXiv:1610.02136},
  year={2016}
}

@article{kim2024vacode,
  title={Vacode: Visual augmented contrastive decoding},
  author={Kim, Sihyeon and Cho, Boryeong and Bae, Sangmin and Ahn, Sumyeong and Yun, Se-Young},
  journal={arXiv preprint arXiv:2408.05337},
  year={2024}
}

@inproceedings{huang2024opera,
  title={Opera: Alleviating hallucination in multi-modal large language models via over-trust penalty and retrospection-allocation},
  author={Huang, Qidong and Dong, Xiaoyi and Zhang, Pan and Wang, Bin and He, Conghui and Wang, Jiaqi and Lin, Dahua and Zhang, Weiming and Yu, Nenghai},
  booktitle={2024 IEEE/CVF Conference on Computer Vision and Pattern Recognition (CVPR)},
  pages={13418--13427},
  year={2024},
  organization={IEEE}
}

@article{park2025second,
  title={Second: Mitigating perceptual hallucination in vision-language models via selective and contrastive decoding},
  author={Park, Woohyeon and Kim, Woojin and Kim, Jaeik and Do, Jaeyoung},
  journal={arXiv preprint arXiv:2506.08391},
  year={2025}
}

\clearpage
\appendix
\section{Implementation Details}
\label{app:impl}

\begin{algorithm}[!b]
\caption{The \method\ correction pipeline.}
\label{alg:magic}
\begin{algorithmic}[1]
\State $\scores \gets \sbase$ \Comment{$1$ forward}
\If{$\margin{\scores} \le \ton$} \Comment{gate the debiasing step}
  \State $\Done \gets \max(\Delta_q,\;\Delta_{q'})$
        \Comment{$\Delta_q = Z(v,q)-Z(v^{\varnothing},q)$}
  \State $\scores \gets \Done$ \Comment{replace with debiased state}
\EndIf
\If{$\margin{\scores} \le \ttwo$} \Comment{gate the restoration step}
  \State $\Dtwo \gets Z(v,q) - Z(\Vsrc,q)$ \Comment{visual residual}
  \If{$\margin{\scores+\Dtwo} - \margin{\scores} \ge \tauV$} \Comment{commit}
    \State $\scores \gets \scores + \Dtwo$ \Comment{restore grounding}
  \EndIf
\EndIf
\State \Return $\arg\max \scores$ \Comment{valid after any gate}
\end{algorithmic}
\end{algorithm}

On Qwen2-VL and LLaVA-NeXT, $\ttwo$ and $\tauV$ are fitted separately for
multiple-choice and open-ended items on the calibration split, since the two
answer formats place the decision margin on different scales, a fixed set of
option letters against a pool of the model's own generations. Based on the
calibration results, we set $\ton$ to $3.25$ and $5.03$ for Qwen2-VL and
LLaVA-NeXT. We set $\ttwo$ to $0.4$ for multiple-choice and $0.5$ for
open-ended items on Qwen2-VL, and to $0.3$ and $0.5$ on LLaVA-NeXT. The
commitment threshold $\tauV$ is $0.625$ for multiple-choice items on both
backbones, and $-0.22$ on Qwen2-VL and $0.375$ on LLaVA-NeXT for open-ended
ones, where the negative value accepts residuals that leave the margin
essentially unchanged. For experiments on the
full distributions, the same procedure lowers $\ton$ to $0.5$ and $2.0$ and
$\ttwo$ to $0.2$, since
most contested samples there are not prior-driven and the gate fires only on
near-ties. The fit replays cached branch scores and costs no additional
model passes. Pool candidates are the model's own short answers, collected
once per sample by greedy decoding and scored by every view. All experiments run
under PyTorch~2.6 on a single NVIDIA A100-80GB GPU.

\section{Proofs and Derivations}

Fix a sample with base scores $\sbase$, let $a=\arg\max\sbase$ and $b$ be the
runner-up, so the margin is $m=\margin{\sbase}=\sbase[a]-\sbase[b]$. A
corrector replaces $\sbase$ by $\scores=\sbase-\est$ with prior estimate
$\est=\prior+\noise$. Because only the top-two gap decides the answer, we
write the signed corrections
\begin{equation}
\dpg \;=\; \prior[a]-\prior[b],\qquad
\de \;=\; \noise[a]-\noise[b],
\label{eq:signed}
\end{equation}
so the correction flips the decision $a\!\to\!b$ if and only if $\dpg+\de>m$.

\subsection{Inertness of a Bounded Correction}
\label{app:law}

\noindent\textbf{Proof of Eq.~\eqref{eq:inert}.} Let
$\scores=\sbase-\est$ with $\|\est\|_\infty\le B$ and let $a,b$ be the top-two
candidates of $\sbase$ with $m=S_{0,a}-S_{0,b}>2B$. Then $\scores_a-\scores_b
= m-(\est_a-\est_b)\ge m-2B>0$, and for any other candidate $c$,
$\scores_a-\scores_c\ge(S_{0,a}-S_{0,c})-2B\ge m-2B>0$. Hence
$\arg\max\scores=a=\arg\max\sbase$ and $\Util(m)=0$ for $m>2B$. The same
argument covers the additive restoration of Eq.~\eqref{eq:commit} with $\Dtwo$
in place of $\est$, and its per-sample form $m_i>2\lVert\Dtwo\rVert_\infty$
holds on $85.1\%$ of the
samples \method\ skips on Qwen2-VL and $94.3\%$ on LLaVA-NeXT. The debiasing
step replaces the state rather than subtracting from it, so its gate is
licensed by the measured utility curve of Figure~\ref{fig:marginlaw} rather
than by this bound. The bound is worst-case, so $\ton$ is fitted on the
calibration set rather than computed from $B$.

\subsection{The Bias Floor and the Branch Budget}
\label{app:budgetproof}

\noindent\textbf{Proof of Eq.~\eqref{eq:budget}.} With
$\est_k-\prior=b_k+\xi_k$ and $\mathbb{E}\,\xi_k=0$, the cross term vanishes
and
\begin{equation}
\mathbb{E}\|\est_k-\prior\|^2=\|b_k\|^2+\mathbb{E}\|\xi_k\|^2
  =\|b_k\|^2+\tfrac{\sigma^2}{k}.
\end{equation}
The first difference is $\Delta_k=(\|b_{k+1}\|^2-\|b_k\|^2)
-\sigma^2/(k(k{+}1))$, so a $(k{+}1)$-th branch lowers the error iff
$\Delta_k<0$, that is iff $\|b_{k+1}\|^2-\|b_k\|^2<\sigma^2/(k(k{+}1))$. The
right side is positive and strictly decreasing in $k$ with limit $0$. Each
perturbation corrupts the input rather than marginalizing the prior,
averaging does not reduce the bias, which the non-zero error floor of
Figure~\ref{fig:biasvar} confirms. The left side is therefore nonnegative.
The error never
falls below $\|b_1\|^2$ and converges to $\|b\|^2$ at constant bias, and the
marginal reduction of an added branch is at most $\sigma^2/(k(k{+}1))$, which
vanishes. The floor and the budget are separate consequences, the first
of the bias and the second of the vanishing return alone. Charging a fixed cost $c$ per
branch on the same scale as the error, the net value of the $(k{+}1)$-th
branch is negative for every $k$ with $\sigma^2/(k(k{+}1))<c$. A finite
optimum $k^\star\approx\sigma/\sqrt{c}$ therefore exists even at $b_k\equiv0$, and its
location is set by the encoder's $\sigma^2$ together with any growth in the
bias increments. That is the model-dependent budget the branch pruning of
\S\ref{sec:prune} estimates. Sweeping the branch count
removes $4.37\%$ of error on Qwen2-VL but only $0.57\%$ on LLaVA-NeXT, so
the variance above the floor is itself encoder-specific.

\section{Additional Analyses}

\subsection{Additional Case Studies}
\label{app:case}

\begin{figure*}[t]\centering
\begin{minipage}[t]{0.485\textwidth}\centering
  {\small\textbf{(a) No gate opens \,---\, $1$ forward pass}}\par\vspace{2pt}
  \casehdr{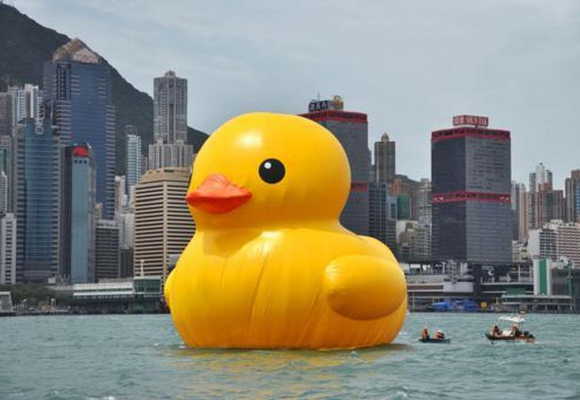}
    {what is the color of this object?}
    {A. green & B. red \\ C. blue & D. yellow}
    {D. yellow}
  \vspace{2pt}
  \casetab{%
   \altrow Base $\sbase$ & $-2.62$ & $-2.12$ & $-1.88$ & $\mathbf{6.62}$ & $8.50$ \\
   Debiased $\Done$ & \multicolumn{4}{c}{gate closed} & \\
   \altrow Residual $\Dtwo$ & \multicolumn{4}{c}{gate closed} & \\
   Final $\scores$ & $-2.62$ & $-2.12$ & $-1.88$ & $\mathbf{6.62}$ & $8.50$ \\}
  \casenote{stgAfill}{stgAline}{\textbf{Base $\to$ D}~\okmark\quad The
    base already separates yellow from every other color by $8.50$. Drawn from
    the natural distribution, where the fitted $\ton$ is $0.5$.}
  \casenoteoff{stgBfill}{stgBline}{\textbf{Step 1: not run}\quad That
    $8.50>\ton{=}0.5$, so the debiasing gate stays shut, no perturbation branch
    is built, and the state is still the one the base pass produced.}
  \casenoteoff{stgCfill}{stgCline}{\textbf{Step 2: not run}\quad The second
    gate reads that same $8.50$, so there is no visual residual either and the
    sample leaves after one pass, the cost of the base model alone.}
\end{minipage}
\hfill
\begin{minipage}[t]{0.485\textwidth}\centering
  {\small\textbf{(b) One gate opens \,---\, $4$ forward passes}}\par\vspace{2pt}
  \casehdr{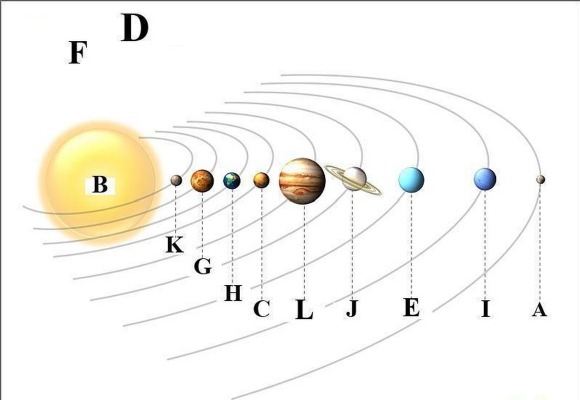}
    {Which planet is depicted by label G?}
    {A. Mars & B. Venus \\ C. Earth & D. Mercury}
    {B. Venus}
  \vspace{2pt}
  \casetab{%
   \altrow Base $\sbase$ & $-3.11$ & $-0.98$ & $-1.86$ & $\mathbf{-0.86}$ & $0.12$ \\
   Debiased $\Done$ & $-0.10$ & $\mathbf{0.65}$ & $0.15$ & $0.06$ & $0.50$ \\
   \altrow Residual $\Dtwo$ & \multicolumn{4}{c}{gate closed} & \\
   Final $\scores$ & $-0.10$ & $\mathbf{0.65}$ & $0.15$ & $0.06$ & $0.50$ \\}
  \casenote{stgAfill}{stgAline}{\textbf{Base $\to$ D}~\xmark\quad Mercury
    leads Venus by $0.12$, the narrowest base margin of the four samples here,
    and $0.12\le\ton{=}3.25$ opens the gate on a barely-made decision.}
  \casenote{stgBfill}{stgBline}{\textbf{Step 1 $\to$ B}~\okmark\quad The
    contrast turns the answer over, from Mercury to Venus, and leaves a margin
    of $0.50$, four times the $0.12$ it started from and enough to settle the
    answer.}
  \casenoteoff{stgCfill}{stgCline}{\textbf{Step 2: not run}\quad That $0.50$
    clears $\ttwo{=}0.4$ by a tenth of a point, so the restoration gate stays
    shut and the sample leaves after $4$ passes, three spent by the first step.}
\end{minipage}
\\[9pt]
\begin{minipage}[t]{0.485\textwidth}\centering
  {\small\textbf{(c) Both gates open, residual commits \,---\, $7$ passes}}\par\vspace{2pt}
  \casehdr{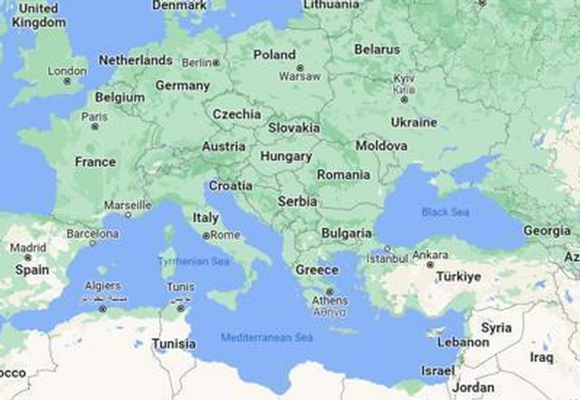}
    {What direction is Serbia in the Mediterranean Sea?}
    {A. east & B. south \\ C. west & D. north}
    {D. north}
  \vspace{2pt}
  \casetab{%
   \altrow Base $\sbase$ & $\mathbf{-0.74}$ & $-2.24$ & $-3.12$ & $-0.99$ & $0.25$ \\
   Debiased $\Done$ & $\mathbf{0.39}$ & $0.30$ & $-1.45$ & $0.05$ & $0.09$ \\
   \altrow Residual $\Dtwo$ & $-0.34$ & $-0.34$ & $-0.97$ & $1.78$ & \\
   Final $\scores$ & $0.05$ & $-0.04$ & $-2.42$ & $\mathbf{1.83}$ & $1.78$ \\}
  \casenote{stgAfill}{stgAline}{\textbf{Base $\to$ A}~\xmark\quad East
    leads north by $0.25$ on a map that puts Serbia well above the sea, and
    that $0.25\le\ton{=}3.25$ opens the debiasing gate on what is close to a
    tie.}
  \casenote{stgBfill}{stgBline}{\textbf{Step 1 $\to$ A}~\xmark\quad
    Debiasing lifts the whole field but not the answer, leaving east ahead by
    $0.09$. That $0.09\le\ttwo{=}0.4$ opens the restoration gate.}
  \casenote{stgCfill}{stgCline}{\textbf{Step 2 $\to$ D}~\okmark\quad The
    residual adds $1.78$ to north alone and moves the rest down. The margin
    widens by $1.69\ge\tauV{=}0.625$, so the step commits.}
\end{minipage}
\hfill
\begin{minipage}[t]{0.485\textwidth}\centering
  {\small\textbf{(d) Both gates open, residual declined \,---\, $7$ passes}}\par\vspace{2pt}
  \casehdr{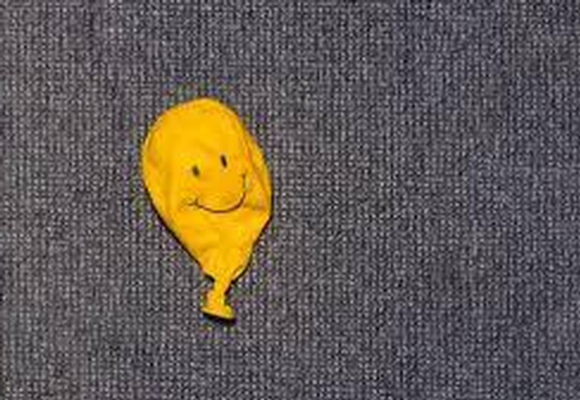}
    {Identify the emotion expressed in this image.}
    {A. sadness & B. anger \\ C. loneliness & D. happiness}
    {D. happiness}
  \vspace{2pt}
  \casetab{%
   \altrow Base $\sbase$ & $-1.98$ & $-6.85$ & $\mathbf{-0.35}$ & $-1.85$ & $1.50$ \\
   Debiased $\Done$ & $0.17$ & $-4.13$ & $0.17$ & $\mathbf{0.37}$ & $0.20$ \\
   \altrow Residual $\Dtwo$ & $1.26$ & $-1.62$ & $-0.24$ & $1.01$ & \\
   Final $\scores$ & $0.17$ & $-4.13$ & $0.17$ & $\mathbf{0.37}$ & $0.20$ \\}
  \casenote{stgAfill}{stgAline}{\textbf{Base $\to$ C}~\xmark\quad
    Loneliness leads happiness by $1.50$, the widest base margin of the three
    biased samples here, and $1.50\le\ton{=}3.25$ opens the debiasing gate.}
  \casenote{stgBfill}{stgBline}{\textbf{Step 1 $\to$ D}~\okmark\quad
    Debiasing already repairs the answer, but only by $0.20$, and that
    $0.20\le\ttwo{=}0.4$ opens the restoration gate on a sample that no longer
    needs it.}
  \casenoteoff{stgCfill}{stgCline}{\textbf{Step 2: rejected}\quad The residual
    would put sadness ahead by $0.05$, narrower than the $0.20$ it starts from.
    That gain of $-0.15$ is below $\tauV{=}0.625$, so the test discards it \okmark.}
\end{minipage}
\caption{Four further Qwen2-VL samples, one per exit of
Algorithm~\ref{alg:magic}.}
\label{fig:case}
\end{figure*}

Figure~\ref{fig:case} takes four samples that Figure~\ref{fig:casebody} does not
use and traces one through each exit of Algorithm~\ref{alg:magic}. In (a) the
base model separates yellow from the other colors by $8.50$, the gate stays
closed, and the sample costs one forward pass. In (b) the debiasing step alone
moves the answer from Mercury to Venus and leaves a margin of $0.50$, so the
restoration step never runs. In (c) that step is not enough on its own, and the
visual residual adds $1.78$ to north and turns the answer over. In (d) the
debiasing step has already repaired the answer and the residual would move it to
sadness on a narrower margin, so the commitment test rejects it and the debiased
answer stands. Panel (a) is drawn from the natural distribution, where the
fitted $\ton$ is $0.5$, because no sample the gate skips on the biased subsets
is answered correctly by the base model.

\subsection{Full Sensitivity Analysis}
\label{app:sensitivity}

Table~\ref{tab:sensitivity-full} reports the same sweep as
Table~\ref{tab:sensitivity} on both backbones. The pattern on LLaVA-NeXT
matches Qwen2-VL: above the fitted value $\ton{=}5.03$, the Overall column is flat, gaining
at most $0.25\%$ as $\ton$ rises to $8$, and the restoration gate $\ttwo$
moves BS Overall by $0.52\%$ while the mean pass count rises from $4.0$ to
$5.5$.

% tab:sensitivity-full -- full two-backbone version of tab:sensitivity, moved
% to the appendix; main paper shows Qwen2-VL only. Same provenance as above.
\begin{table}[!t]
\centering
\renewcommand{\arraystretch}{1.0}
\setlength{\tabcolsep}{5pt}
\resizebox{\columnwidth}{!}{%
\begin{tabular}{l c cccc}
\Xhline{1.2pt}
\headrow Model & Value & MCQ & Others & Overall & Passes \\
\Xhline{1.2pt}
\multicolumn{6}{l}{\emph{Debiasing gate $t_1$}} \\
\altrow & 0.5 & 18.81 & 29.30 & 21.84 & 3.0 \\
 & 1.0 & 24.68 & 31.64 & 26.69 & 3.7 \\
\altrow & 2.0 & 31.03 & 34.38 & 32.00 & 4.4 \\
 & 3.0 & 32.30 & 34.96 & 33.07 & 4.8 \\
\altrow & 5.0 & 32.38 & 35.74 & 33.35 & 5.0 \\
 \multirow{-6}{*}{Qwen2-VL} & 8.0 & \textbf{32.46} & \textbf{35.94} & \textbf{33.47} & 5.0 \\
\Xhline{0.6pt}
\altrow & 0.5 & 21.36 & 45.73 & 27.31 & 2.7 \\
 & 1.0 & 25.05 & 47.11 & 30.43 & 3.3 \\
\altrow & 2.0 & 29.23 & 49.75 & 34.23 & 4.0 \\
 & 3.0 & 30.48 & 51.76 & 35.67 & 4.4 \\
\altrow & 5.0 & 31.13 & 52.01 & 36.22 & 4.7 \\
 \multirow{-6}{*}{LLaVA-NeXT} & 8.0 & \textbf{31.29} & \textbf{52.51} & \textbf{36.47} & 4.8 \\
\Xhline{1.2pt}
\multicolumn{6}{l}{\emph{Restoration gate $t_2$}} \\
\altrow & 0.0 & 30.32 & 33.59 & 31.26 & 4.0 \\
 & 0.2 & 31.35 & 33.40 & 31.94 & 4.4 \\
\altrow & 0.3 & 32.14 & 33.59 & 32.56 & 4.6 \\
 & 0.4 & 32.30 & 34.18 & 32.84 & 4.9 \\
\altrow & 0.6 & \textbf{32.46} & \textbf{35.16} & \textbf{33.24} & 5.1 \\
 \multirow{-6}{*}{Qwen2-VL} & 1.0 & 32.06 & \textbf{35.16} & 32.96 & 5.5 \\
\Xhline{0.6pt}
\altrow & 0.0 & 30.48 & 52.39 & 35.83 & 4.0 \\
 & 0.2 & 30.89 & 52.39 & 36.13 & 4.5 \\
\altrow & 0.3 & 31.13 & \textbf{52.51} & \textbf{36.35} & 4.7 \\
 & 0.4 & 31.17 & 52.26 & 36.32 & 4.9 \\
\altrow & 0.6 & 31.25 & 51.88 & 36.29 & 5.1 \\
 \multirow{-6}{*}{LLaVA-NeXT} & 1.0 & \textbf{31.29} & 51.76 & 36.29 & 5.5 \\
\Xhline{1.2pt}
\end{tabular}}
\caption{Sensitivity to the two allocation thresholds on the BS Subset of
DRBench, on both backbones. We bold the best result in each column within a
block.}
\label{tab:sensitivity-full}
\end{table}

\subsection{Cost Analysis}
\label{app:cost}

A fired step costs what a perturbation branch
costs, so cost is set by how often the gate fires rather than by a cheaper
branch. On MMStar, where SCI reports its own overhead, the gate fires on about
a quarter of the samples, and
\method\ spends $1.98$ passes on Qwen2-VL and
$3.47$ on LLaVA-NeXT against $\text{SCI}_5$'s five. On the biased subsets
almost every sample is contested, and Table~\ref{tab:cost} reports the pass
counts there. A scoring pass costs $915$\,ms on Qwen2-VL and $130$\,ms on
LLaVA-NeXT, and a generation pass $1339$\,ms and $473$\,ms, measured on one
A100-80GB at the deployed image budget. At those pass counts \method\ takes
$4.57$\,s per sample on Qwen2-VL against $4.58$\,s for $\text{SCI}_5$ and
$6.41$\,s for $\text{SCI}_7$, and $0.69$\,s on LLaVA-NeXT against $0.65$\,s
and $0.91$\,s.

% tab:cost -- average inference passes per sample (measured, not nominal).
\begin{table}[t]
\centering
\renewcommand{\arraystretch}{1.15}
\resizebox{\columnwidth}{!}{%
\begin{tabular}{l cc c cc}
\Xhline{1.2pt}
\headrow & \multicolumn{2}{c}{Biased (BS)} && \multicolumn{2}{c}{Natural} \\
\cline{2-3}\cline{5-6}
\headrow \multirow{-2}{*}{Method} & MCQ & Others && MCQ & Others \\
\Xhline{1.2pt}
\altrow SCI$_5$ & 5.0 & 5.0 && 5.0 & 5.0 \\
SCI$_7$ & 7.0 & 7.0 && 7.0 & 7.0 \\
\altrow\method\ (Qwen2-VL)   & 5.0 & 4.5 && 1.5 & 2.8 \\
\method\ (LLaVA-NeXT) & 4.9 & 4.2 && 2.4 & 2.5 \\
\Xhline{1.2pt}
\end{tabular}}
\caption{Average inference passes per sample.}
\label{tab:cost}
\end{table}

\subsection{Perturbation Invariance}
\label{app:manifold}

\begin{figure}[t]\centering
\includegraphics[width=\columnwidth]{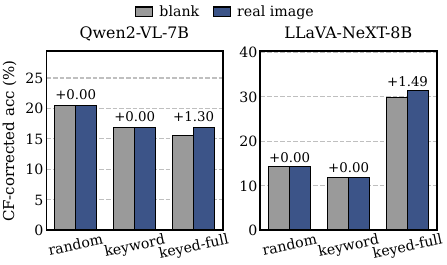}
\caption{Random and keyword-matched replacements decide as a blank image
does. The keyed-full condition answers the question and is the positive
control.}
\label{fig:manifold}
\end{figure}

Several recent methods are built on the assumption that perturbing the image
puts the branch off-distribution, whether by moving the counterfactual inside
the forward pass \citep{qin2026sira} or by replacing the perturbed branch with
a cross-image one \citep{zhao2025cicd}. At the decision point we do not
reproduce that premise, as Figure~\ref{fig:manifold} shows. We replace the image
with another real photograph rather than a corrupted one, drawn at random or
retrieved by keyword from the question, and the branch still decides as it does
on a blank image. What we measure is invariance at the decision rather than in
the distribution, over replacements that do not answer the question.

\subsection{Attribution Analysis}
\label{app:attribution}

We examine whether the gain comes from the margin gate or from the two-step
correction it fires. Table~\ref{tab:family} compares \method\ against the two
adaptive CD methods nearest to it, which already vary their correction per
sample, and against a corrector that keeps our gate and nothing else. Neither
VACoDe \citep{kim2024vacode} nor CHASD \citep{huang2026chasd} beats \method\ on
any cell. Keeping our gate but dropping the second contrast and the residual
costs $3.0\%$ in BS Overall on Qwen2-VL and $1.1\%$ on LLaVA-NeXT, so the gate alone
does not carry the gain either. Two further checks agree. Adding our gate to
SCI's fusion leaves every prediction on the calibration set unchanged, and
replacing the margin with maximum softmax probability \citep{hendrycks2017msp}
or predictive entropy, at a matched firing rate, moves BS Overall by at most
$0.08\%$. What the gate measures therefore matters little, and the separation
comes from the correction it fires.

% tab:family -- answer-level adaptations of the two nearest gated/adaptive
% contrast methods, replayed from the frozen step-0 dumps.
% Provenance: research/malt_family_baselines.py (PREREG in
% research/log/malt_family_baselines/); test reads disclosed in App.
\begin{table}[t]
\centering
\renewcommand{\arraystretch}{1.15}
\resizebox{\columnwidth}{!}{%
\begin{tabular}{l cc c cc}
\Xhline{1.2pt}
\headrow & \multicolumn{2}{c}{Qwen2-VL} && \multicolumn{2}{c}{LLaVA-NeXT} \\
\cline{2-3}\cline{5-6}
\headrow \multirow{-2}{*}{Method} & S & BS && S & BS \\
\Xhline{1.2pt}
\altrow VACoDe-A (max contrast) & 46.91 & 22.28 && \textbf{40.99} & 26.78 \\
VACoDe-A (min contrast) & 38.55 & 32.20 && 31.09 & 29.38 \\
\altrow CHASD-A (uncertainty-gated) & \textbf{48.73} & 30.53 && 40.30 & 30.47 \\
gate only (blank contrast) & 45.35 & 29.29 && 39.52 & 30.00 \\
\altrow \method & \textbf{48.73} & \textbf{32.30} && \textbf{40.99} & \textbf{31.13} \\
\Xhline{1.2pt}
\end{tabular}}
\caption{Our answer-level adaptations of the nearest adaptive-contrast
methods, MCQ cells on the test split.}
\label{tab:family}
\end{table}

Both baselines act token-by-token during generation, so each port keeps its
selection rule and re-reads it over the candidate scores on the
multiple-choice test split. VACoDe picks the corruption whose answer softmax
is farthest from the original's and feeds it to the VCD operator, as its
published rule prescribes. CHASD rebuilds its counterfactual from the regions
its attention map marks as salient, which has no analogue over a fixed
candidate set. Our port therefore drops that step and gates a noised-image
contrast by predictive entropy, in place of its published maximum-probability
threshold. The margin-gate-only row keeps our gate over a single blank contrast
and drops the restoration step. Free constants are fitted on the calibration
set \method\ uses, over a grid that extends in the baseline-favoring
direction.

\begin{table*}[t]
\centering
\renewcommand{\arraystretch}{1.15}
\begin{tabular}{p{0.44\textwidth} p{0.40\textwidth}}
\Xhline{1.2pt}
\headrow Instruction issued by the benchmark & Answer-format rephrasing \\
\Xhline{1.2pt}
\altrow Please select the correct answer from the options above.
  & Return only the final option letter, with no explanation. \\
Please answer yes or no. & Return exactly one word: yes or no. \\
\altrow Answer the question using a single word or phrase.
  & Return only the final short answer, without a full sentence. \\
Answer the question directly. & Return only the final answer, without a full
  sentence. \\
\Xhline{1.2pt}
\end{tabular}
\caption{The answer-format rephrasings the debiasing step contrasts against,
one representative benchmark instruction per row. Variant wordings of the
same instruction map to the same rephrasing.}
\label{tab:prompts}
\end{table*}

\section{Prompt Templates}

Table~\ref{tab:prompts} lists the rephrasing pairs, matched on the
instruction the benchmark itself issues. 

\end{document}